\pdfoutput=1

\documentclass[11pt]{article}

\usepackage[preprint]{acl}

\usepackage{times}
\usepackage{latexsym}
\usepackage{amsmath}
\usepackage{makecell}
\usepackage{changepage}

\usepackage[T1]{fontenc}

\usepackage[utf8]{inputenc}
\usepackage{multirow}
\usepackage{booktabs}

\usepackage{inconsolata}

\usepackage{graphicx}

\usepackage{algorithm}
\usepackage{algpseudocode}

\usepackage{colortbl}

\newcommand{\shade}[1]{\ifnum\numexpr#1<0 0\else#1\fi}

\def\ccadoit[#1]#2{\cellcolor{WildStrawberry!\the\numexpr#1/2.5}{ #2}}

\newcommand{\cellcolortoken}[1]{%
    \begingroup
    \ifnum#1<75
        \cellcolor{WildStrawberry!5}#1
    \else\ifnum#1<80
        \cellcolor{WildStrawberry!10}#1
    \else\ifnum#1<85
        \cellcolor{WildStrawberry!15}#1
    \else\ifnum#1<90
        \cellcolor{WildStrawberry!20}#1
    \else\ifnum#1<95
        \cellcolor{WildStrawberry!25}#1
    \else\ifnum#1<98
        \cellcolor{WildStrawberry!30}#1
    \else
        \cellcolor{WildStrawberry!35}#1
    \fi
    \endgroup
}
\title{Stepwise Perplexity-Guided Refinement for Efficient Chain-of-Thought Reasoning in Large Language Models}

\begin{document}
\author{\\
\textbf{~~~~ ~~ ~~ Yingqian Cui$^{1,2}$\thanks{{Work done during her internship at Amazon.}}~~ Pengfei He$^{1}$ ~~ Jingying Zeng$^{2}$ ~~ Hui Liu$^{2}$ ~~ Xianfeng Tang$^{2}$ ~~ } \\
\textbf{ Zhenwei Dai$^{2}$ ~~Yan Han$^{2}$ ~~ Chen Luo$^{2}$ ~~ Jing Huang$^{2}$ ~~ } \\
\textbf{Zhen Li$^{2}$ ~~ Suhang Wang$^{3}$~~ Yue Xing$^{1}$ ~~ Jiliang Tang$^{1}$ ~~ Qi He$^{2}$ }\\
 ~~ $^{1}$Michigan State University
 ~~$^{2}$ Amazon ~~ $^{3}$ Pennsylvania State University \\
\\ 
\\
}
\maketitle
\begin{abstract}
Chain-of-Thought (CoT) reasoning, which breaks down complex tasks into intermediate reasoning steps, has significantly enhanced the performance of large language models (LLMs) on challenging tasks. However, the detailed reasoning process in CoT often incurs long generation times and high computational costs, partly due to the inclusion of unnecessary steps. To address this, we propose a method to identify critical reasoning steps using perplexity as a measure of their importance: a step is deemed critical if its removal causes a significant increase in perplexity. Our method enables models to focus solely on generating these critical steps. This can be achieved through two approaches: refining demonstration examples in few-shot CoT or fine-tuning the model using selected examples that include only critical steps. Comprehensive experiments validate the effectiveness of our method, which achieves a better balance between the reasoning accuracy and efficiency of CoT.
\end{abstract}

\section{Introduction}\label{sec:intro}
Large language models (LLMs) are powerful generative models capable of performing diverse tasks in different domains~\cite{gramopadhye2024few,karabacak2023embracing,ling2024deductive} and demonstrating strong reasoning capabilities \cite{jaech2024openai}. Recent advancements, such as few-shot/zero-shot Chain-of-Thought (CoT) \cite{wei2022chain,kojima2022large}, as well as fine-tuning \cite{liu2023improving}, have significantly enhanced the LLMs' reasoning capabilities by leveraging intermediate reasoning steps.
In particular, through few-shot CoT, LLMs can learn from the reasoning steps in the demonstration examples and apply similar reasoning patterns to target tasks. In the case of zero-shot CoT, LLMs are prompted to"think step by step" to generate reasoning steps. In fine-tuning, LLMs can also learn from the reasoning steps in the fine-tuning samples, further enhancing their reasoning abilities.

While many existing reasoning methods rely on available data (e.g., few-shot examples or fine-tuning datasets), there is limited understanding of which reasoning steps are truly essential and how their impact varies across different models. 
This gap hinders progress in two key areas: (1) how to effectively identify and remove unimportant reasoning steps from the data to reduce computational costs, and (2)
whether the important reasoning steps for one model are also important to another.

\begin{figure*}
    \centering
      \vspace{-0.3in}
    \includegraphics[width=0.8\linewidth]{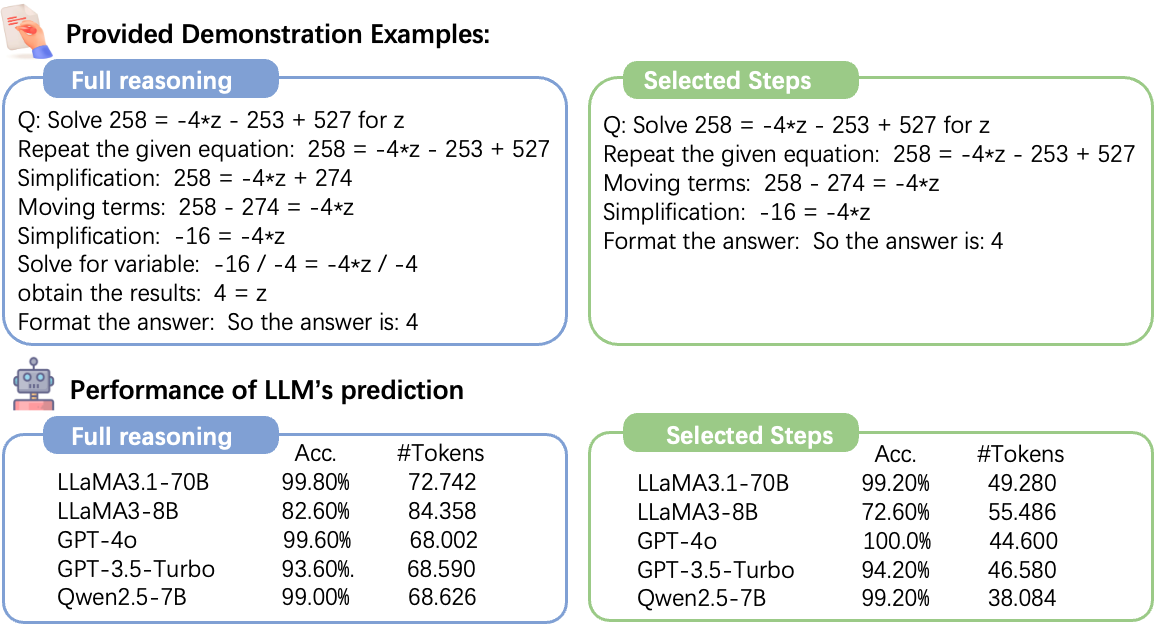}
    \vspace{-0.1in}
    \caption{\small Prediction accuracy of few-shot CoT using all/selected steps in the demonstration examples.}
    \label{fig:intro}
    \vspace{-0.15in}
\end{figure*}
For example, we observe that removing certain reasoning steps from the demonstrations in few-shot CoT can have varying effects: some models {follow the modified examples and} generate much fewer tokens while maintaining reasoning accuracy, whereas others experience a decline in performance. Specifically, we consider a math problem of function solving \cite{saxton2019analysing}.
{We compare two versions of demonstrations when conducting few-shot CoT: one with full {manually crafted} reasoning paths and another containing only intuitively important steps, as shown in Figure~\ref{fig:intro}. For most models, removing certain steps significantly reduces the number of generated tokens with minimal impact on accuracy, suggesting that the removed steps contribute limited meaningful information. However, LLaMA3-8B shows a noticeable decline in accuracy, indicating that the importance of reasoning steps can vary across different LLMs.

Similar to the few-shot CoT scenario, when given a set of fine-tuning samples with reasoning steps, some LLMs may find some steps redundant, and the fine-tuning cannot improve the prediction accuracy. However, these LLMs will follow the fine-tuning samples to generate the additional tokens, raising the computation cost. Other LLMs may struggle to develop reasoning capabilities when given too few reasoning steps during fine-tuning. This observation will be further demonstrated in Section~\ref{sec:experiment}.

Therefore, in this work, we focus on identifying unimportant reasoning steps from few-shot examples or fine-tuning data given a specific LLM. To achieve this, we propose a method leveraging perplexity, a metric commonly used to measure the confidence or fluency of model-generated text~\cite{jelinek1977perplexity}, to quantify the impact of each reasoning step.
Our contributions are as follows:  

First, since perplexity reflects an LLM’s confidence in processing inputs and generating outputs~\cite{jelinek1977perplexity}, we hypothesize that perplexity can serve as an indicator of reasoning step importance. Specifically, if the perplexity changes significantly after removing a reasoning step, we conjecture that the removed step plays a crucial role in the model’s decision-making process. To validate this hypothesis, we conduct empirical analyses (Section~\ref{sec:pre-corr}) and observe a strong correlation between changes in perplexity (with and without a reasoning step) and the prediction performance. This finding reveals that perplexity effectively quantifies the significance of individual reasoning steps.

Second, inspired by this insight, we develop an algorithm, \textbf{S}tepwise \textbf{P}erplexity-Gu\textbf{I}ded \textbf{R}ef\textbf{I}nemen\textbf{T} (SPIRIT), to remove or merge unimportant reasoning steps. 
To effectively apply this approach across different scenarios of CoT, we tailor our approach for two different use cases,  (1) few-shot CoT, where the full reasoning steps in the examples are known (SPIRIT-FS), and (2) fine-tuning, where the samples only have input and the final answer at the beginning (SPIRIT-FT). 

{When developing the algorithms}, a common technical challenge is that some steps, though considered unimportant by the selection criteria, may still contain partial usefulness. Removing such steps could disrupt the coherence of the remaining reasoning process.  To address this, we further refine the algorithm by incorporating a merging mechanism to ensure the overall coherence of the whole reasoning process.

Finally, we conduct comprehensive experiments to examine the effectiveness of the proposed algorithms.  
In few-shot CoT, our method successfully provides demonstrations that guide the model to generate a more efficient reasoning process without {greatly} sacrificing performance. For fine-tuning, our approach achieves a better effectiveness-efficiency trade-off than randomly select steps to be removed.

\vspace{-0.03in}
\section{Preliminary}\label{sec:prelim}
\vspace{-0.02in}
In this section, we first present the essentials of perplexity, and then introduce our exploration on how to use perplexity to analyze the reasoning steps.
\subsection{Perplexity (PPL)}
 Perplexity was developed in \cite{jelinek1977perplexity} and is a common metric for LLMs. It is defined as
{\small\begin{align}
 &\text{PPL}(x,\{w_k\}_{k=1}^{N}) \hspace{2.5in}\nonumber\\
& =\quad\exp\left(-\frac{1}{N} \sum_{i=1}^{N} \log p(w_i \mid x, w_1,\dots, w_{i-1})\right), \label{eq:ppl}
\end{align}} 
\vspace{-0.05in}

\noindent where $x$ represents the prompt, 
{$\{w_k\}_{k=1}^{N}$ denotes sequence of tokens with total length $N$ which are conditioned on $x$.} The probability \( p( w_i \mid x,  w_1, w_2, \dots, w_{i-1}) \) is the likelihood assigned by the model to the \( i \)-th token given the prompt and the preceding tokens.

In literature, many studies utilize perplexity, e.g., for reference model pruning~\cite{ankner2024perplexed},  attack detection~\cite{alon2023detecting}, misinformation detection~\cite{lee2020misinformation}, and uncertainty quantification~\cite{cooper2024perplexed}.

\subsection{Relationship between Perplexity and CoT Prediction Accuracy}\label{sec:pre-corr}
We conduct preliminary evaluation to investigate the relationship between PPL and CoT prediction accuracy when changing the steps used in the reasoning procedure. Intuitively, a higher likelihood indicates that the LLM is more confident to the context, and from Eq.(\ref{eq:ppl}), a higher likelihood results is a lower PPL. Thus, we hypothesize that the PPL is negatively correlated with the prediction accuracy.

In the experiments summarized in Table~\ref{tab:correlation}, we apply few-shot demonstrations to perform CoT reasoning across three tasks from the DeepMind Mathematics Dataset~\cite{saxton2019analysing}: Solving linear equation (AL1), calculating derivative (Diff-Calc), and measuring time difference (Time-Diff). 
For each dataset, we manually construct the demonstration examples. All the constructed examples in the same dataset share the same reasoning steps. Then we randomly select steps to be removed from all examples in demonstration and calculate the perplexity of the resulting generation and the accuracy of CoT reasoning.
Table~\ref{tab:correlation} presents the correlation coefficient between the perplexity and accuracy and the p-value indicating the statistical significance of their negative relationship.
Notably, the perplexity for all experiments is computed using LLaMA3-7B, while accuracy is assessed based on generations from both LLaMA3-7B and GPT-4o-mini (in a transfer case). 

The results from Table \ref{tab:correlation} indicate a statistically significant negative correlation between perplexity and accuracy across all tasks, aligning with our hypothesis. This observation paves us a way to identify unimportant reasoning steps from the reasoning path: Since the correlation is negative, if we remove some steps while maintaining the perplexity of the sample, then it is likely that there will be no accuracy loss, i.e., the removed steps are unimportant. Furthermore, the correlation appears transferable across models, as perplexity computed with LLaMA3-7B remains strongly correlated with accuracy evaluated using GPT-4o-mini, indicating the potential transferability of our proposed method.

\begin{table}[t]
\centering \label{tab:corr}
\renewcommand{\arraystretch}{1.2}
\caption{\small Correlation Between Perplexity of Reasoning Generation and Reasoning Accuracy, with p-Values Indicating Statistical Confidence}
  \vspace{-0.1in}
\label{tab:correlation}
\resizebox{0.45\textwidth}{!}{
\begin{tabular}{lcc|cc}
\toprule
& \multicolumn{2}{c|}{\textbf{LLaMA3-8B}} & \multicolumn{2}{c}{\textbf{GPT-4o-mini}} \\
\cmidrule{2-5}
& \textbf{r} & \textbf{p-value} & \textbf{r} & \textbf{p-value} \\
\midrule
AL1 & {-0.690} & 0.0272 & {-0.860} & 0.0014 \\
Diff-Calc & -0.997 & $3.37e{-8}$ & -0.993 & $4.88e{-7}$ \\
Time-Diff &{-0.850} & 0.0154 & {-0.973} & 0.0002 \\
\bottomrule
\end{tabular}}
\vspace{-0.2in}
\end{table}
  \vspace{-0.07in}
\section{The Proposed Algorithm - SPIRIT}\label{sec:method}
  \vspace{-0.06in}
In this section, we present the details of SPIRIT. Since few-shot CoT and fine-tuning {utilize data in different ways}, 
we first provide the general idea in Section~\ref{sec:general} and then describe case-specific details in Section~\ref{sec:few_shot} (\textbf{F}ew-\textbf{S}hot CoT, SPIRIT-FS) and \ref{sec:fine-tuning} (\textbf{F}ine-\textbf{T}uning, SPIRIT-FT), respectively.
\vspace{-0.05in}
\subsection{General Idea}\label{sec:general}
\vspace{-0.04in}
For both few-shot CoT and fine-tuning, the general idea is to select unimportant reasoning steps and then process them. 
When removing one reasoning step, the final PPL will be changed. We enumerate all reasoning steps to get the one whose removal results in the lowest PPL.

On the other hand, a concern with step removal is that directly eliminating a step from a structured reasoning process can lead to coherence issues, particularly when the step contains intermediate results necessary for subsequent computations. For example, consider the reasoning process in Figure~\ref{fig:merge}. If we remove the step "So, the number of students present is 40 - 4 = 36 students.", the value 36 appears abruptly in the following step "36 * 3/4 = 27"
without proper context, making the solution difficult to follow. In such cases, merging steps is necessary to maintain coherence. An appropriate revision could be "(40-4)*3/4 = 27". 
Based on these observations, we propose to incorporate a merging paradigm into the algorithm, whose details will be introduced in the following subsections.

\vspace{-0.06in}
\subsection{Few-Shot CoT (SPIRIT-FS)}\label{sec:few_shot}
\vspace{-0.04in}
\begin{algorithm*}[!ht]
\caption{SPIRIT-FS}\label{alg:few_shot}
\begin{algorithmic}[1]
\State \textbf{Input}: Demonstration set $\mathcal{D} = \{(q^d_i, \mathcal{R}_i)\}$, calibration set $\mathcal{C} = \{q^c_i\}_{i=1}^m$, threshold $t$
\State Initialize $\mathcal{D}^* \leftarrow \mathcal{D}$
\While{True}
    \State Find the most unimportant step $j^* \leftarrow \arg\min_j \frac{1}{m} \sum_{i} \text{PPL}(\{\mathcal{D^*} \backslash r^j, q^c_i\}, \mathcal{M}(\mathcal{D^*} \backslash r^j, q^c_i))$
    \State Update perplexity $\text{PPL}_{\text{best}} \leftarrow \frac{1}{m} \sum_{i} \text{PPL}(\{\mathcal{D^*} \backslash r^{j^*}, q^c_i\},\mathcal{M}(\{\mathcal{D^*} \backslash r^{j^*}, q^c_i\}))$
    \State {Derive merged reasoning $\mathcal{D^*}_{\text{merge}}$, ensuring coherence}
    \State {\textbf{if} removal step limit reached \textbf{then break else}}    $\mathcal{D^*} \leftarrow \mathcal{D^*}_{\text{merge}}$
\EndWhile
\State \Return Refined demonstration $\mathcal{D}^*$
\end{algorithmic}
\end{algorithm*}

When performing few-shot CoT, we assume the demonstration examples follow a consistent reasoning format, e.g., for the function solving problem, all examples follow the same steps as in Figure~\ref{fig:intro}. For simplicity, we treat one sentence as one step in the algorithm. Our goal is to remove unimportant reasoning steps in the predefined demonstration examples.

The detailed procedure of SPIRIT-FS is outlined in Algorithm~\ref{alg:few_shot}. For a demonstration set $\mathcal{D} = \{(q^d_i, \mathcal{R}_i)\}$, $q^d_i$ represents a demonstration question and $\mathcal{R}_i=(r_i^1,r_i^2,\ldots)$ denotes its corresponding reasoning process with the reasoning steps $r_i^1,r_i^2,\ldots$. 
{The calibration set $\mathcal{C} = \{q^c_i\}$ is a set of questions from the dataset, containing tens of examples, used to assess the impact of reasoning step removal by evaluating perplexity changes.}
We iteratively refine $\mathcal{D}$ by removing unnecessary reasoning steps. 
At each iteration, we evaluate the impact of removing each step $r^j$ by computing the average of $\text{PPL}(\{\mathcal{D} \backslash r^j, q^c_i\}, \mathcal{M}(\{\mathcal{D} \backslash r^j, q^c_i\}))$ over the calibration set ($\mathcal{M}(\cdot)$ denotes the LLM and $A\backslash b$ means removing the element b from set A). The step $r^{j^*}$ that minimizes the perplexity will be pruned {for all demonstration examples}. 

{To maintain coherence, instead of direct removal,  step $r_i^{j^*}$ is merged with other steps, using either an LLM or human effort, in a way as the example shown in Figure~\ref{fig:merge}. 
The merging process integrates the step with either the preceding or subsequent step, depending on the semantic meaning to ensure coherence. If an LLM is used for merging, we provide demonstration examples in the prompt to guide the process. This procedure is repeated until the stopping criteria is met, e.g., a specified number of steps to be removed (used in our few-shot CoT experiments), or a perplexity threshold (used in fine-tuning experiments).}} 

  \vspace{-0.05in}
\subsection{Fine-Tuning (SPIRIT-FT)}\label{sec:fine-tuning}
  \vspace{-0.03in}
  The full details of SPIRIT-FT are presented in Algorithm~\ref{alg:fine-tune}.
Compared to few-shot CoT, some changes are made for the fine-tuning scenario. 

First, in fine-tuning, {not all datasets contains complete reasoning steps.} 
For datasets with high-quality annotated reasoning steps, we directly use the provided reasoning. However, for datasets that only include rationales or lack explicit reasoning step, we employ a capable LLM, such as GPT-4o or LLaMA3.1-70B, to generate the {the full reasoning steps} based on the input and final answer.
After obtaining the reasoning steps, we apply Algorithm \ref{alg:fine-tune} to refine them.

Second, {due to the different scenario of few-shot CoT and fine-tuning, }the perplexity calculation is handled differently:
In few-shot CoT, given the prompt, we compute $\text{PPL}(\{\mathcal{D}, q^c_i\}, \mathcal{M}(\{\mathcal{D}, q^c_i\}))$, the perplexity based on the actual model generation in inference. {We use a calibration set to compute the average perplexity over calibration examples, guiding the refinement of reasoning steps. The refined steps are then applied to new testing examples.} {In contrast, in fine-tuning, when refining the reasoning steps, we do not have access to inference-time perplexity after fine-tuning. The perplexity in this case is calculated directly on the fine-tuning data, i.e., $\text{PPL}(q_i, \mathcal{R}_i)$. There is no calibration set involved, as the step selection is performed on the fine-tuning data itself rather than requiring a separate set for evaluation.} 

\begin{figure}[h]
  \vspace{-0.02in}
    \centering 
    \includegraphics[width=1\linewidth]{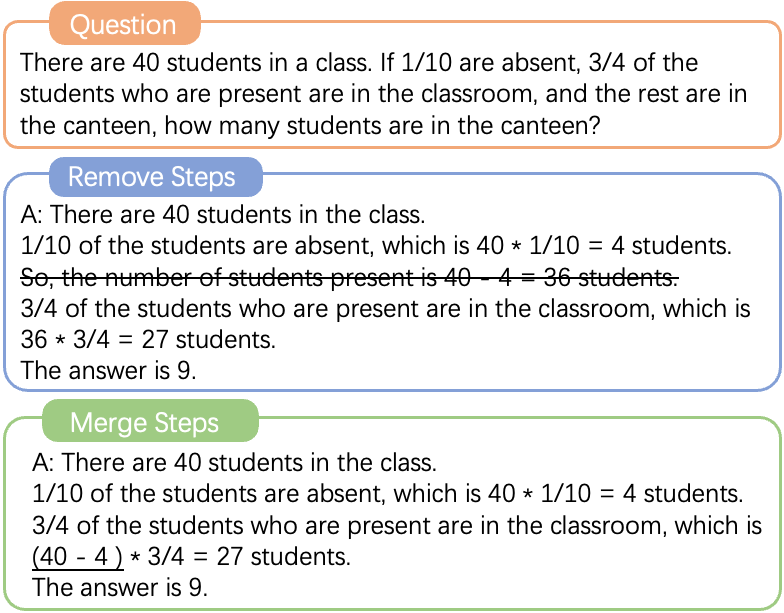}    \vspace{-0.25in}
    \caption{\small Comparison of removing and merging.}   \label{fig:merge}
    \vspace{-0.1in}
\end{figure}
\begin{algorithm*}[!ht]
\vspace{-0.05in}
\caption{SPIRIT-FT}
\label{alg:fine-tune}
\begin{algorithmic}[1]
\State \textbf{Input}: Questions $\mathcal{Q} = \{q_i\}$, reasoning processes $\mathcal{R} = \{\mathcal{R}_i\}$, thresholds $t_1, t_2$
\For{each sample $i$} 
    \State Initialize $\mathcal{R}_i^* \leftarrow \mathcal{R}_i$, $\text{PPL}_{\text{orig}} \leftarrow \text{PPL}(q_i, \mathcal{R}_i^*)$
    \While{True}
        \State Get the most unimportant step $r_{\text{worst}} \leftarrow \arg\min_{r_j \in \mathcal{R}_i^*} \text{PPL}(q_i, \mathcal{R}_i^* \backslash \{r_j\})$
        \State Update perplexity $\text{PPL}_{\text{rem}} \leftarrow \text{PPL}(q_i, \mathcal{R}_i^* \backslash \{r_{\text{worst}}\})$
        \If{$\text{PPL}_{\text{rem}} > t_2 \cdot \text{PPL}_{\text{orig}}$} \textbf{break}
        \ElsIf{$\text{PPL}_{\text{rem}} < t_1 \cdot \text{PPL}_{\text{orig}}$}  $  \mathcal{R}_i^*\leftarrow  \{ \mathcal{R}_i^*\backslash r_{\text{worst}}\}$
        \Else
            \State Generate merged reasoning $\mathcal{R}_{\text{merge}}$, ensuring coherence
            \State  $\mathcal{R}_i^* \leftarrow\mathcal{R}_{\text{merge}}$ if $\text{PPL}(q_i, \mathcal{R}_{\text{merge}}) < \text{PPL}_{\text{rem}}$, else  $  \mathcal{R}_i^*\leftarrow  \{ \mathcal{R}_i^*\backslash r_{\text{worst}}\}$
        \EndIf
    \EndWhile
\EndFor
\State \Return Refined reasoning processes $\mathcal{R}^* = \{\mathcal{R}_i^*\}$
\end{algorithmic}
\end{algorithm*}

To explain the details of Algorithm \ref{alg:fine-tune}\footnote{{Although Algorithm \ref{alg:few_shot} allows different ways for merging and stopping, in the fine-tuning scenario, to handle the large amount of fine-tuning data and the diversity of the reasoning steps among the data, we explicitly design the merging and stopping criteria for SPIRIT-FT.}}, given a set of questions $\mathcal{Q} = \{q_i\}$ and their corresponding reasoning processes $\mathcal{R} = \{\mathcal{R}_i\}$, we iteratively refine each reasoning process $\mathcal{R}_i$, 
by selectively removing or merging reasoning steps. At each iteration, we identify the step $r_{\text{worst}}$ whose removal minimizes perplexity $\text{PPL}(q_i, \mathcal{R}_{i^*} \backslash {r_i^j})$. If the resulting perplexity $\text{PPL}_{\text{rem}}$ falls below a threshold $t_1$ relative to the original perplexity, the step is directly removed. Otherwise, we generate a merged version of the reasoning process and compare its perplexity $\text{PPL}_{\text{merge}}$ with $\text{PPL}_{\text{rem}}$, selecting the option with the lower perplexity. 
This process continues iteratively until the resulting perplexity exceeds a threshold $t_2$, at which point refinement is terminated. 

{We apply capable LLMs to conduct the merging. The  merging prompt (include several examples) can be found in Appendix~\ref{sec:appd:prompt}. To save computation cost, we do not merge steps when  $\text{PPL}_{\text{rem}}$ is below $t_1$. To justify this design, we provide experiment results (in {Appendix \ref{sec:appd:t1}}) to demonstrate that it is more necessary to conduct merging when $\text{PPL}_{\text{rem}}$ is large. In contrast, for small $\text{PPL}_{\text{rem}}$, merging provides only trivial improvement.

  \vspace{-0.03in}
\section{Experiment}\label{sec:experiment}
  \vspace{-0.03in}
In this section, we conduct comprehensive experiments to demonstrate the effectiveness of SPIRIT. We present the results of SPIRIT-FS in Section~\ref{sec:exp-s} and demonstrate the performance of SPIRIT-FT in Section~\ref{sec:exp-f}. Both sections include the discussion on the transferability of SPIRIT by investigating whether the reasoning step selection process generalizes across different models. Due to page limit, we postpone 
the ablation studies in Appendix~\ref{sec:ablation}, where we examine the impact of some key components in the design of SPIRIT-FT.

  \vspace{-0.02in}
\subsection{Few-shot CoT (SPIRIT-FS)}\label{sec:exp-s}
  \vspace{-0.02in}

\noindent\textbf{Datasets.} We consider the Algebra-Linear-1d  Task (AL1) and Number-Base-Conversion Task (NBC) from the Mathematics Dataset~\cite{saxton2019analysing} for the experiments. For both tasks we randomly select 500 examples for evaluation. 

\noindent\textbf{Language Models.} Our experiments use five LLMs: GPT-3.5-Turbo~\cite{brown2020language}, GPT-4o-mini~\cite{brown2020language}, LLaMA3-8B-Instruct, LLaMA3.1-70B-Instruct~\cite{grattafiori2024llama3} and Qwen2.5-7B-Instruct~\cite{qwen2024qwen2.5} (LLaMA3-8B, LLaMA3.1-70B, Qwen2.5-7B in short). The temperature is set to 0 to ensure deterministic outputs in generation. Notably, when applying our algorithm to open-source models (LLaMA3-8B, LLaMA3.1-70B, and Qwen2.5-7B), we use the corresponding model to compute perplexity and refine the reasoning demonstrations. For GPT-4o-mini and GPT-3.5-Turbo, where direct perplexity computation is unavailable, we instead use LLaMA3.1-70B to estimate perplexity and generate the refined demonstration examples (in a transfer case). We show details of the hyperparameters of fine-tuning in Appendix~\ref{sec:appd:hyper}.
\begin{table*}[h]
\vspace{-0.3in}
\caption{\small Performance of using Algorithm 1 for steps selection in few-shot CoT with Algebra-linear-1d task.}\label{tab:equation}
\vspace{-0.1in}
\centering
\resizebox{0.93\textwidth}{!}{
\begin{tabular}{lccc|cc|cc|cc|cc} 
\toprule
\multirow{2}{*}{Method} & & \multicolumn{2}{c|}{LLaMA3.1-70B} & \multicolumn{2}{c|}{LLaMA3-8B} & \multicolumn{2}{c|}{Qwen 2.5} & \multicolumn{2}{c|}{GPT-3.5-Turbo} & \multicolumn{2}{c}{GPT-4o-mini} \\ 
\cmidrule{2-12}
 & & acc(\%) & \# tokens & acc(\%) & \# tokens & acc(\%) & \# tokens & acc(\%) & \# tokens & acc(\%) & \# tokens \\ 
\midrule
Zero-shot & & \cellcolor{WildStrawberry!60} 99.60 & \cellcolor{CornflowerBlue!5} 134.186 & \cellcolor{WildStrawberry!60} 86.40 & \cellcolor{CornflowerBlue!5} 115.698 & \cellcolor{WildStrawberry!60} 99.60 & \cellcolor{CornflowerBlue!5} 142.418 & \cellcolor{WildStrawberry!15} 87.60 & \cellcolor{CornflowerBlue!5} 97.474 & \cellcolor{WildStrawberry!60} 99.00 & \cellcolor{CornflowerBlue!5} 191.104 \\ 
\midrule
\makecell{Few-shot}&(7 steps) & \cellcolor{WildStrawberry!60} 99.80 & \cellcolor{CornflowerBlue!15} 72.742 & \cellcolor{WildStrawberry!35} 82.00 & \cellcolor{CornflowerBlue!15} 84.358 & \cellcolor{WildStrawberry!60} 99.00 & \cellcolor{CornflowerBlue!15} 68.626 & \cellcolor{WildStrawberry!60} 93.60 & \cellcolor{CornflowerBlue!15} 68.59 & \cellcolor{WildStrawberry!35} 98.00 & \cellcolor{CornflowerBlue!15} 66.95 \\

\midrule
\multirow{3}{*}{\makecell{Few-shot\\ (4 steps)}} 
 & Ours (remove) & \cellcolor{WildStrawberry!35} 99.20 & \cellcolor{CornflowerBlue!35} 49.28 & \cellcolor{WildStrawberry!35} 72.60 & \cellcolor{CornflowerBlue!35} 55.486 & \cellcolor{WildStrawberry!60} 99.20 & \cellcolor{CornflowerBlue!60} 38.084 & \cellcolor{WildStrawberry!60} 94.20 & \cellcolor{CornflowerBlue!15} 46.58 & \cellcolor{WildStrawberry!35} 98.40 & \cellcolor{CornflowerBlue!35} 47.43 \\

 & Ours (merge)  &\cellcolor{WildStrawberry!35} 99.20 &
\cellcolor{CornflowerBlue!35} 55.478  & 
\cellcolor{WildStrawberry!35} 71.40  & 
 \cellcolor{CornflowerBlue!35}55.814 & \cellcolor{WildStrawberry!35}
 97.80	&
 \cellcolor{CornflowerBlue!35} 41.78	&
 \cellcolor{WildStrawberry!35} 91.63&	 \cellcolor{CornflowerBlue!15}
 49.185&	
 \cellcolor{WildStrawberry!60}98.80	& \cellcolor{CornflowerBlue!35} 
 49.40 \\ 
 & Rand & \cellcolor{WildStrawberry!15} 94.80 & \cellcolor{CornflowerBlue!35} 48.01 & \cellcolor{WildStrawberry!5} 57.00 & \cellcolor{CornflowerBlue!35} 51.892 & \cellcolor{WildStrawberry!15} 93.60 & \cellcolor{CornflowerBlue!35} 46.726 & \cellcolor{WildStrawberry!5} 84.60 & \cellcolor{CornflowerBlue!35} 42.363 & \cellcolor{WildStrawberry!15} 94.40 & \cellcolor{CornflowerBlue!35} 41.34 \\

\midrule
\multirow{3}{*}{\makecell{Few-shot\\ (3 steps)}} 
 & Ours (remove) & \cellcolor{WildStrawberry!15} 95.60 & \cellcolor{CornflowerBlue!60} 35.934 & \cellcolor{WildStrawberry!15} 62.00 & \cellcolor{CornflowerBlue!60} 42.86 & \cellcolor{WildStrawberry!15} 95.40 & \cellcolor{CornflowerBlue!60} 35.938 & \cellcolor{WildStrawberry!35} 91.40 & \cellcolor{CornflowerBlue!60} 34.536 & \cellcolor{WildStrawberry!15} 97.00 & \cellcolor{CornflowerBlue!60} 34.196 \\
 
 & Ours (merge)  &\cellcolor{WildStrawberry!15}96.20 &	\cellcolor{CornflowerBlue!35}50.894	&
\cellcolor{WildStrawberry!15} 63.2&
\cellcolor{CornflowerBlue!35} 44.792	&\cellcolor{WildStrawberry!35}
 97.00&	
  \cellcolor{CornflowerBlue!35} 40.614&	
  \cellcolor{WildStrawberry!35} 90.93&   \cellcolor{CornflowerBlue!35}
 38.074&	
 \cellcolor{WildStrawberry!15} 96.80	&
   \cellcolor{CornflowerBlue!35} 36.824\\ 
 & Rand & \cellcolor{WildStrawberry!5} 80.40 & \cellcolor{CornflowerBlue!35} 41.576 & \cellcolor{WildStrawberry!5} 59.00 & \cellcolor{CornflowerBlue!35} 50.00 & \cellcolor{WildStrawberry!5} 86.80 &  \cellcolor{CornflowerBlue!35} 41.768 & \cellcolor{WildStrawberry!5} 82.40 & \cellcolor{CornflowerBlue!35} 37.188 & \cellcolor{WildStrawberry!5} 78.60 & \cellcolor{CornflowerBlue!35} 37.2 \\
 
\midrule
Concise& & \cellcolor{WildStrawberry!35} 98.40 & \cellcolor{CornflowerBlue!15} 77.038 & \cellcolor{WildStrawberry!15} 64.60 & \cellcolor{CornflowerBlue!15} 66.276 & \cellcolor{WildStrawberry!35} 97.40 &\cellcolor{CornflowerBlue!15}58.874& \cellcolor{WildStrawberry!15} 85.40 & \cellcolor{CornflowerBlue!15} 54.39 & \cellcolor{WildStrawberry!15} 96.80 & \cellcolor{CornflowerBlue!35} 36.82 \\
 
\bottomrule
\end{tabular}
}
\end{table*}
\begin{table*}[h]
\vspace{-0.1in}
\caption{\small Performance of using Algorithm 1 for steps selection in few-shot CoT with Number-Base-Conversion task.}\label{tab:base}
\vspace{-0.1in}
\centering
\resizebox{0.93\textwidth}{!}{
\begin{tabular}{lccc|cc|cc|cc|cc} 
\toprule
\multirow{2}{*}{Method} & & \multicolumn{2}{c|}{LLaMA3.1-70B} & \multicolumn{2}{c|}{LLaMA3-8B} & \multicolumn{2}{c|}{Qwen 2.5} & \multicolumn{2}{c|}{GPT-3.5-Turbo} & \multicolumn{2}{c}{GPT-4o-mini} \\ 
\cmidrule{2-12}
 & & acc(\%) & \# tokens & acc(\%) & \# tokens & acc(\%) & \# tokens & acc(\%) & \# tokens & acc(\%) & \# tokens \\ 
\midrule
 Zero-shot & & \cellcolor{WildStrawberry!5} 75.40 & \cellcolor{CornflowerBlue!5} 244.10 & \cellcolor{WildStrawberry!5} 36.40 & \cellcolor{CornflowerBlue!5} 195.00 &\cellcolor{WildStrawberry!35 } 82.80 & \cellcolor{CornflowerBlue!5}272.99 & \cellcolor{WildStrawberry!5 }62.00&	\cellcolor{CornflowerBlue!5}166.39 & \cellcolor{WildStrawberry!15 }  92.63 & \cellcolor{CornflowerBlue!5} 319.74 \\ 
\midrule
Few-shot& (12 steps)& \cellcolor{WildStrawberry!60 }95.60 & \cellcolor{CornflowerBlue!15}147.12 & \cellcolor{WildStrawberry!60 } 62.40 &\cellcolor{CornflowerBlue!5}  151.77 & \cellcolor{WildStrawberry!60 } 88.60 & \cellcolor{CornflowerBlue!15} 157.43 &\cellcolor{WildStrawberry!60 }  84.20 &\cellcolor{CornflowerBlue!5}  	161.24 & \cellcolor{WildStrawberry!35 }  95.80 &\cellcolor{CornflowerBlue!5} 156.66 \\ 
\midrule
\multirow{3}{*}{\makecell{Few-shot \\ (9 steps)}} 
 & Ours (remove)  &\cellcolor{WildStrawberry!60 } 95.00 &\cellcolor{CornflowerBlue!35} 107.29 & \cellcolor{WildStrawberry!35 }59.40    &\cellcolor{CornflowerBlue!15}  122.67 & \cellcolor{WildStrawberry!35 }  84.20&	\cellcolor{CornflowerBlue!15}128.69 & \cellcolor{WildStrawberry!60 } 85.40 &	\cellcolor{CornflowerBlue!35} 113.09 & \cellcolor{WildStrawberry!60 } 97.00 & \cellcolor{CornflowerBlue!35} 120.28 \\ 
  & Ours  (merge) &\cellcolor{WildStrawberry!60 } 94.40 & \cellcolor{CornflowerBlue!35} 110.66  &\cellcolor{WildStrawberry!60 }  60.00 &\cellcolor{CornflowerBlue!15}  132.24 & \cellcolor{WildStrawberry!35 } 85.60&\cellcolor{CornflowerBlue!15} 	129.87  & \cellcolor{WildStrawberry!60 } 86.80 &\cellcolor{CornflowerBlue!15} 118.85 & \cellcolor{WildStrawberry!60 } 97.80 & \cellcolor{CornflowerBlue!15} 124.68\\ 
 & Rand  & \cellcolor{WildStrawberry!15 }86.60 & \cellcolor{CornflowerBlue!15}114.46 &\cellcolor{WildStrawberry!35 } 52.40 & \cellcolor{CornflowerBlue!35} 117.69 &\cellcolor{WildStrawberry!15 }  80.60 &\cellcolor{CornflowerBlue!35} 123.23 &\cellcolor{WildStrawberry!15 }  72.00 &\cellcolor{CornflowerBlue!15}	122.26&\cellcolor{WildStrawberry!15 }  91.60 & \cellcolor{CornflowerBlue!15}  137.28 \\ 
\midrule
\multirow{3}{*}{\makecell{Few-shot \\ (6 steps)}} 
 & Ours (remove) & 89.20\cellcolor{WildStrawberry!35 } & \cellcolor{CornflowerBlue!60} 92.51 &\cellcolor{WildStrawberry!15 } 44.60 & \cellcolor{CornflowerBlue!60} 93.27 &\cellcolor{WildStrawberry!15 }   75.40 &\cellcolor{CornflowerBlue!60}	91.28  & \cellcolor{WildStrawberry!35 }  77.80 &	\cellcolor{CornflowerBlue!60} 97.41 & \cellcolor{WildStrawberry!15 } 93.00 & \cellcolor{CornflowerBlue!60} 106.93 \\ 
   & Ours  (merge) &\cellcolor{WildStrawberry!35 }90.60 &  \cellcolor{CornflowerBlue!60} 95.73 & \cellcolor{WildStrawberry!15 }49.80 & \cellcolor{CornflowerBlue!35}  104.39 &\cellcolor{WildStrawberry!15 }  77.80 &	\cellcolor{CornflowerBlue!35} 97.66  &\cellcolor{WildStrawberry!35 }  79.40 & \cellcolor{CornflowerBlue!60} 103.21 &\cellcolor{WildStrawberry!35 }  96.60 & \cellcolor{CornflowerBlue!60} 108.52\\ 
 & Rand  & \cellcolor{WildStrawberry!15 } 81.60 & \cellcolor{CornflowerBlue!15} 117.99 &\cellcolor{WildStrawberry!5 }  41.80 & \cellcolor{CornflowerBlue!60} 101.35 & \cellcolor{WildStrawberry!5 }  63.40 &	\cellcolor{CornflowerBlue!60}92.57 &\cellcolor{WildStrawberry!5 }  69.20 &	\cellcolor{CornflowerBlue!35} 115.60&\cellcolor{WildStrawberry!15 }   86.40 &\cellcolor{CornflowerBlue!15}  129.52 \\ 
\midrule
Concise & & \cellcolor{WildStrawberry!5 }73.60 & \cellcolor{CornflowerBlue!35} 111.65 &\cellcolor{WildStrawberry!15 } 44.00 & \cellcolor{CornflowerBlue!60} 100.51 &\cellcolor{WildStrawberry!15 }  77.00 &\cellcolor{CornflowerBlue!5} 	161.80 &\cellcolor{WildStrawberry!5 } 58.80 &	\cellcolor{CornflowerBlue!35} 115.14 & \cellcolor{WildStrawberry!5 }  72.60 & \cellcolor{CornflowerBlue!35} 112.64 \\ 
\bottomrule
\end{tabular}}
\vspace{-0.15in}
\end{table*}

\noindent\textbf{Procedures.}
For both AL1 and NBC, we manually create the detailed reasoning solution for the demonstration examples and apply SPIRIT-FS to refine the reasoning paths. For AL1, we reduce the reasoning process from 7 steps to 3 or 4 steps.
For NBC, we reduce the reasoning from 12 steps to 9 or 6 steps. We present the corresponding accuracy of few-shot CoT in Table~\ref{tab:equation} and~\ref{tab:base}, labeled as "Ours (merge)". To measure the efficiency, we show the number of generated tokens. 
To validate the effectiveness of SPIRIT-FS, we compare the performance with two baselines methods, (1) randomly select steps to be removed ("Rand"); and (2) directly ask the model to be concise in generation ("Concise"). Additionally, we include another variant of our method, labeled as "Ours (remove)", where we refine reasoning steps using SPIRIT-FS but apply only removal without merging.

\noindent\textbf{Results.} From the results in Table~\ref{tab:equation} and~\ref{tab:base}, it is observed that in general, across different models and tasks, our algorithm achieves a better trade-off between accuracy and efficiency by maintaining higher accuracy under a similar number of generated tokens. For example, except for LLaMA3-8B, all other models maintain comparable accuracy  when the number of reasoning steps is reduced from 7 to 4 in the AL1 task. Similarly, in the NBC task, performance remains stable when steps are reduced from 12 to 9, except for LLaMA3-8B and Qwen 2.5-7B, which experience a slight drop in accuracy.
In contrast, baseline methods "Concise" and 
"Rand" tend to sacrifice much more accuracy when the reasoning length is reduced.

In addition, comparing "Ours (merge)" and "Ours (removal)", it is observed that for the simpler AL1 task, merging does not yield a significant accuracy improvement, while slightly increasing the number of generated tokens. But for the more difficult task NBC, "Ours (merge)" demonstrate a better accuracy, indicating the necessity of merging to ensure performance in more complex reasoning scenarios.

\noindent\textbf{Transferability.} 
{From the results in Table~\ref{tab:equation} and~\ref{tab:base}, we can see that, reasoning step selection based on the perplexity of LLaMA3.1-70B leads to good performance when applied to GPT-4o-mini and GPT-3.5-turbo. Specifically, for the AL1 and NBC tasks, when the number of reasoning steps is reduced to 4 and 9, respectively, accuracies remain unchanged or even slightly improve. As steps are further reduced, accuracies gradually decrease, but still outperforms both random step removal and the approach of simply prompting the model to be more concise. This suggests that perplexity-based step selection generalizes well across models.}

\vspace{-0.05in}
\subsection{Fine-Tuning (SPIRIT-FT)}\label{sec:exp-f}
\vspace{-0.03in}
\noindent\textbf{Datasets.} 
We consider two main datasets including GSM8K~\cite{cobbe2021training} and MetaMathQA~\cite{yu2023metamath}. For GSM8K, the entire training set (with 7.4k examples) is utilized for example refinement and fine-tuning, with evaluation performed on the full evaluation set (with 1.3k examples). For MetaMathQA, we randomly select 19k examples for refinement and fine-tuning, while 1.95k examples are selected as the testing data.

\noindent\textbf{Language Models.} Our main experiments involve two LLMs: LLaMA3-8B-Instruct and Qwen2.5-7B-Instruct (LLaMA3-8B, Qwen2.5-7B in short).

\noindent\textbf{Fine-tuning Methods. } 
We consider two fine-tuning methods including Supervised Fine-tuning (SFT) and Odds Ratio Preference Optimization (ORPO)~\cite{hong2024orpo}.
We applied LoRA~\cite{hu2021lora} for both methods.

\noindent\textbf{Procedures.}
We applied SPIRIT-FT to refine the reasoning paths, fine-tuned the model with the refined data, and evaluated the fine-tuned model by measuring both prediction accuracy and the number of generated tokens. The trade-off between accuracy and efficiency was controlled by adjusting $t_2$, which determines the extent of step removal/merging. Notably, when fine-tuning with different models, we used the specific model itself to compute perplexity for unimportant step determination. We present the relationship between accuracy and efficiency across different models and different datasets in Figure~\ref{fig:sft} and~\ref{fig:orpo} for SFT and  ORPO, respectively. The results are labeled as "Min PPL (merge)". 

{For evaluation}, in the experiments of SFT, we compare {SPIRIT-FT} with three control sets, (1) a variant of {SPIRIT-FT} where we {only remove but not merge steps}
("Min PPL (remove)"); 
(2) randomly select steps to be removed ("Randomly remove"); and 
(3) applying an inverse of Algorithm~\ref{alg:fine-tune} to {remove the most important steps whose removal maximize the perplexity}
("Max PPL (Remove)"). For ORPO, we utilize some of the above datasets to form chosen/rejected pairs: (1) Chosen: Min PPL (Merge) / Rejected: Max PPL (Remove);
(2) Chosen: Min PPL (Remove) / Rejected: Max PPL (Remove);
(3) Chosen: Max PPL (Remove)/ Rejected: Min PPL (Remove). The labels for the above settings are "Min PPL (merge)", "Min PPL (remove)" and "Max PPL (remove)", respectively.
\begin{figure}[h]
    \centering
    \includegraphics[width=1.02\linewidth]{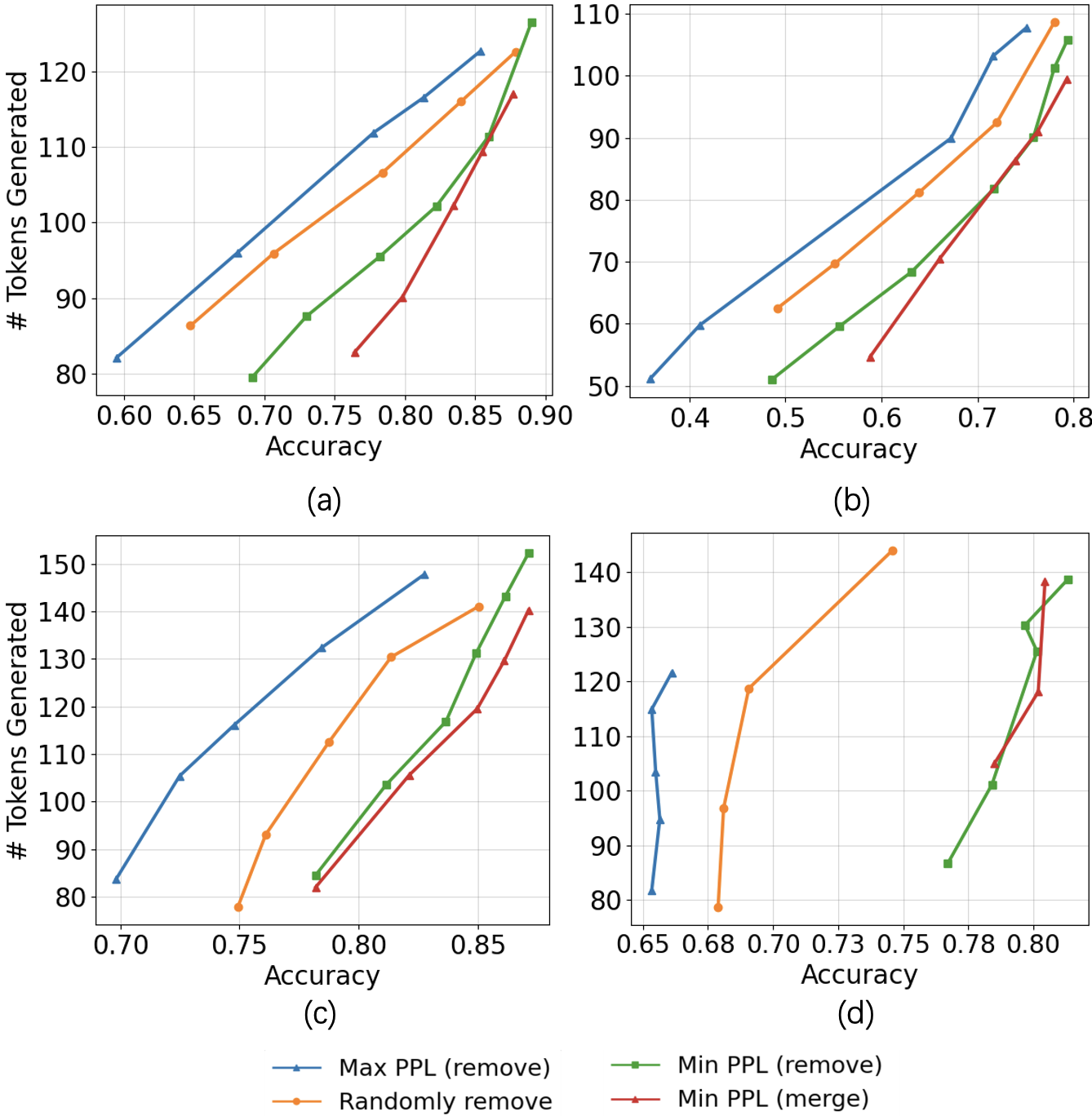}
 \vspace{-0.2in}
    \caption{\small Accuracy-Efficiency Relation when fine-tuning with SFT. (a) Qwen2.5-7B, GSM8K; (b) LLAMA3-8B, GSM8K; (c) Qwen2.5-7B, MetaMathQA; (b) LLAMA3-8B, MetaMathQA }
    \label{fig:sft}
    \vspace{-0.1in}
\end{figure}
\begin{figure}[h]
    \centering
    \includegraphics[width=1.02\linewidth]{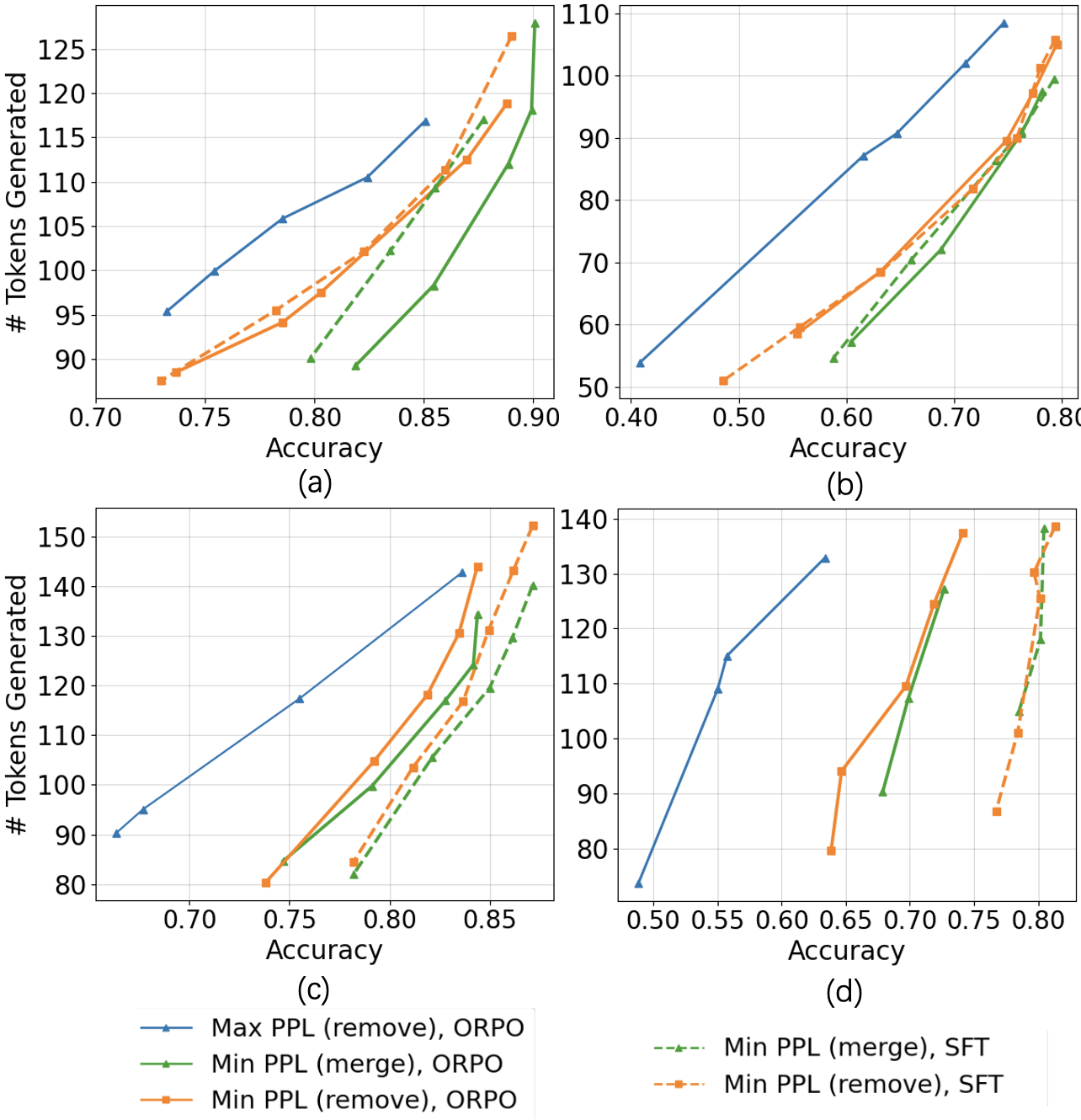}
    \vspace{-0.2in}
    \caption{\small Accuracy-Efficiency Relation when fine-tuning with ORPO. (a) Qwen2.5-7B, GSM8K; (b) LLAMA3-8B, GSM8K; (c) Qwen2.5-7B, MetaMathQA; (b) LLAMA3-8B, MetaMathQA}
    \label{fig:orpo}
    \vspace{-0.2in}
\end{figure}

\noindent\textbf{Results.}
Based on the SFT results in Figure~\ref{fig:sft}, 
across different models and datasets, compared with randomly selecting steps to be removed, SPIRIT-FT consistently demonstrate a better trade-off between accuracy and efficiency by achieving a higher accuracy when the number of generated tokens is similar. In addition, the performance of "Randomly remove" is better than "Max PPL (remove)", which provide further evidence that perplexity is {effective in measuring the importance of the reasoning steps.}
Comparing the results of "Min PPL (remove)" and "Min (merge)", 
the algorithm with merging demonstrates a better performance than directly removing steps, which confirms the necessity of conducting merging to maintain coherence in the reasoning process. 

For the results regarding ORPO in Figure~\ref{fig:orpo}, a general order of the performance among different sets {in terms of accuracy-efficiency trade-offs is} "Min PPL (merge)" > "Min PPL (remove)" > "Max PPL (remove)". These results also provide evidence that minimizing perplexity is an effective criterion for selecting reasoning steps, and incorporating merging further enhances performance by preserving coherence in the reasoning process. 

\noindent \textbf{Transferability.}
\label{sec:exp-transfer}
We examine the transferability of SPIRIT-FT across models in Figure~\ref{fig:transfer}. 
It shows the results where LLaMA3-8B is used to calculate perplexity, and the refined dataset is subsequently applied to fine-tune either LLaMA2-7B-Chat or Qwen1.5-7B-Chat.
For comparison, we also provide the results in which the step removal is performed using the perplexity computed by the same model as the fine-tuning target. 

From Figure~\ref{fig:transfer} we can see that, in general, the {the {ranking} of the performance among} "Max PPL (Remove)," "Randomly Remove," "Min PPL (Remove)," and "Min PPL (Merge)" remain consistent even when the perplexity is computed using a different model. This suggests that the LLaMA3-8B exhibit similar patterns with LLaMA2-7B and Qwen2.5-7B in how to process and learn from data, indicating a shared understanding of reasoning step importance {and a transferability of perplexity across models}. 

{On the other hand}, a surprising observation in Figure~\ref{fig:transfer} is that when applying the method to LLaMA2-7B and Qwen1.5-7B, using the perplexity of LLaMA3-7B to calculat perplexity results in even better prediction performance than using the corresponding LLMs themselves for determining unimportant steps. To explain this, our conjecture is that the perplexity of weaker LLMs is influenced by additional factors beyond the true importance of reasoning steps {such as the coherence as a human language (i.e., utility \citep{shi2024muse}) and the understanding of math notations \cite{zhang2024mathverse}}, making it less effective for uncertainty quantification {for the reasoning itself}.
\begin{figure}[h]
\vspace{-0.15in}
    \centering
    \includegraphics[width=0.83\linewidth]{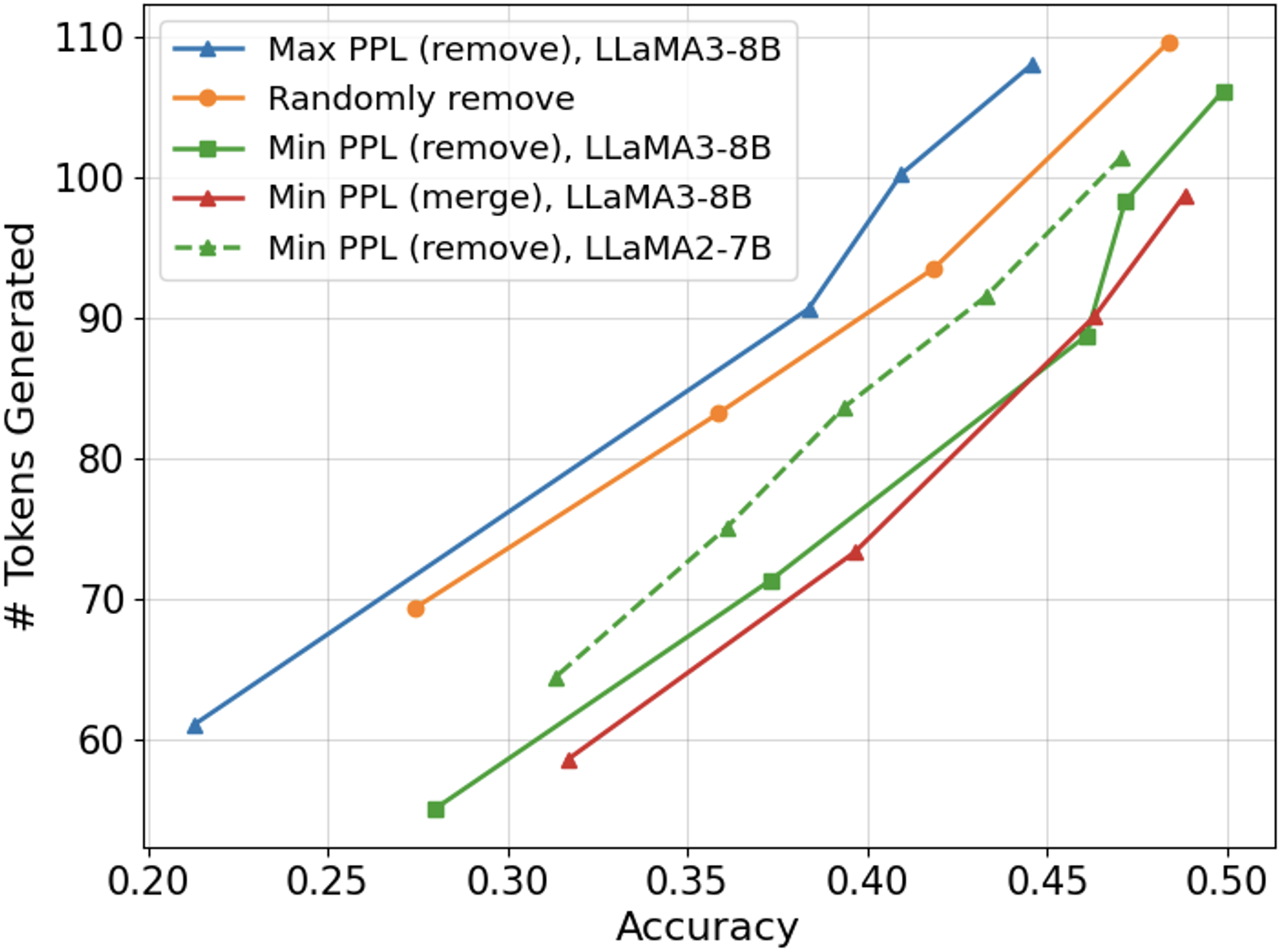}
        \includegraphics[width=0.83\linewidth]{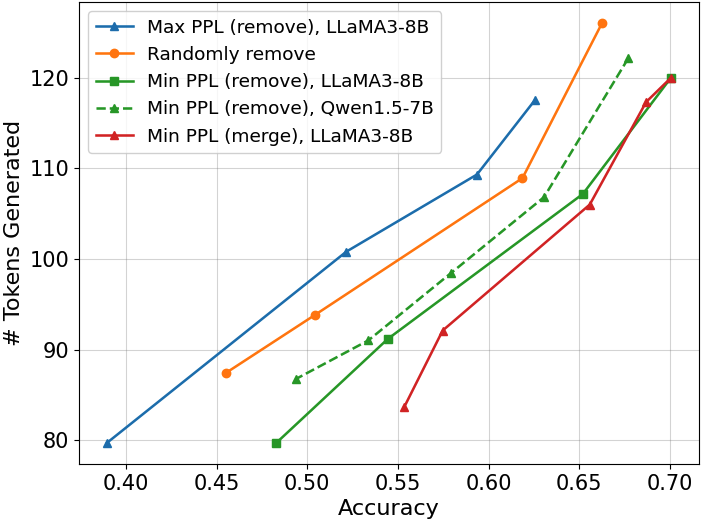}
          \vspace{-0.1in}
    \caption{\small Transferability of PPL when calculated using LLaMA3-8B and evaluated on LLaMA2-7B / Qwen1.5-7B.}
    \label{fig:transfer}
    \vspace{-0.15in}
\end{figure}
\vspace{-0.03in}
\section{Related Works}\label{sec:related}
\vspace{-0.0in}
\noindent\textbf{Inference-Stage Techniques in LLM Reasoning.} Many studies aim to enhance LLM reasoning at the inference stage, without modifying model weights. Early work \cite{wei2022chain} uses few-shot demonstrations to guide reasoning, while \cite{kojima2022large} shows that simply prompting the LLM to "think step by step" also improves the accuracy without demonstrations. Subsequent techniques, such as Graph-of-Thoughts \cite{besta2024graph}, Tree-of-Thoughts \cite{yao2024tree}, and Forest of Thoughts \cite{bi2024forest}, further adapt the reasoning paradigm. Other works focus on self-consistency \cite{wang2022self,wan2023better} or structured input analysis \cite{he2024make}. Different from the aforementioned literature, our work examines the importance of each reasoning step.


\noindent\textbf{CoT Fine-Tuning.}
In literature and real practice, there are two common types of LLM fine-tuning methods: supervised fine-tuning (SFT) and reinforcement learning (RL)-based alignment methods.

SFT is commonly used to adapt an LLM to downstream task, and various studies have investigated SFT. For example, \cite{zhou2024lima} hypothesizes that LLMs require only a few samples from the target task to align with desired behaviors.\cite{dong2023abilities} explores how SFT affects different LLM capabilities, while \cite{ovadia2023fine} compares fine-tuning with retrieval-augmented generation, and \cite{ling2024deductive} investigates overfitting in SFT. Other works focus on data selection for SFT, such as \cite{shen2024rethinking} and \cite{zhang2024balancing}.

RL-based alignment methods incorporate preference labels into loss function, e.g., reinforcement learning with human feedback \cite{ziegler2019fine}, direct preference optimization (DPO) \cite{rafailov2024direct}, ORPO \cite{hong2024orpo}, BCO \cite{jung2024binary}, and KTO \cite{ethayarajhmodel}.
  \vspace{-0.08in}
\section{Conclusion}
  \vspace{-0.08in}
In this paper, we introduce SPIRIT, a method for refining reasoning steps in few-shot CoT and CoT fine-tuning for improving reasoning efficiency while maintaining accuracy. Based on the observation that changes in perplexity correlate with reasoning step importance, SPIRIT works by iteratively identifying unimportant steps {through evaluating the change in perplexity,}
then merge the unimportant steps. 
Experiments demonstrate the effectiveness of SPIRIT in improving the trade-off between accuracy and efficiency in both few-shot CoT and CoT in fine-tuning.


\section*{Limitations}

{While the main observation in Section \ref{sec:exp-transfer} is on the transferability of the algorithm, we also observe that the perplexity from the stronger model (LLaMA3-8B) works even better than using the weaker model's own perplexity (Qwen1.5-7B and LLaMA2-7B) in selecting the unimportant reasoning steps. This implies that perplexity contains more information than what is needed in SPIRIT, indicating the potential limitation of using perplexity in the algorithm: If we want to fine-tune an even weaker model, we would better use a stronger model's perplexity. 
This observation also implies the potential interplay between data quality and the model's capability: A "good" quality with high-quality complex reasoning steps may not benefit a weak model. We believe this observation can inspire future works in data attrition and data selection to consider the model's own capability. 

Another limitation is that, in the algorithm and experiments, we assume reasoning steps among few-shot examples match with each other sentence by sentence. This can be further enhanced if the reasoning steps match the general pattern. However, since different tasks have diverse reasoning patterns, we anticipate that such an enhancement should be specifically designed for the given task and dataset.

}

\bibliography{custom}

\newpage
\onecolumn
\appendix
\label{sec:appendix}
\section{Ablation Studies}\label{sec:ablation}
In this section, we conduct ablation studies by applying different variations of SPIRIT-FT to validate the reasonableness behind the key components in the design of the algorithm:\\

\noindent(1) \textbf{Always applying merging and no removal:} 
Instead of comparing the effects of merging and removal, we modify the approach to always apply merging after selecting a step for refinement. \\
(2) \textbf{Removing the threshold $t_1$:}
, meaning that after determining which step to remove, we no longer check if the resulting perplexity is below a threshold. Instead, we always proceed with merging and then compare the effects of merging versus removal.\\
The results of (1) and (2) are presented as scatter point in Figure~\ref{fig:abl-merge} and~\ref{fig:abl-t1} respectively, labeled as "Always merging" or "Removing $t_1$ threshold", respectively, with comparisons to the performance of the original algorithm.

From the results in Figure~\ref{fig:abl-merge}, we observe that always applying merging leads to performance comparable to the original algorithm, when the number of generated tokens is high. However, as the number of tokens is reduced below 80, performance degrades significantly compared to the original design, indicating that blindly merging steps without considering removal can compromise reasoning effectiveness. 

In addition, Figure~\ref{fig:abl-t1} shows that, when removing $t_1$ threshold, performance appears to improve slightly. However, this comes at the cost of greatly increased computation, as the algorithm involves more rounds of merging. This results highlight that our method provides a more computationally efficient approach while effectively preserving performance.
\begin{figure}[h]
    \centering
    \includegraphics[width=0.5\linewidth]{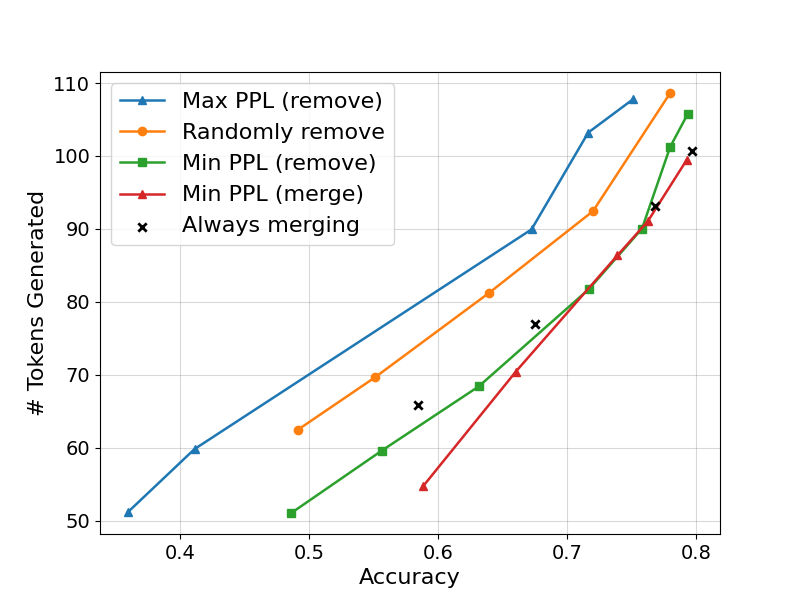}
    \caption{\small Performance of SPIRIT-FT when always applying merging}
    \label{fig:abl-merge}
\end{figure}

\begin{figure}[h]
    \centering
    \includegraphics[width=0.5\linewidth]{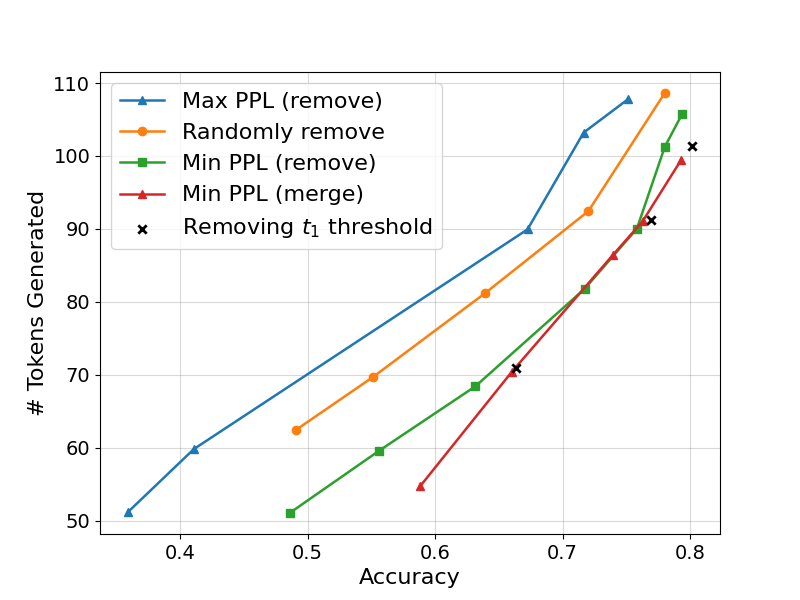}
    \caption{\small Performance of SPIRIT-FT when removing the $t_1$ threshold.}
    \label{fig:abl-t1}
\end{figure}

\section{Additional Implementation Details and Adjustments}
\noindent\textbf{Perplexity Calculation Adjustment}
   In practice, when calculating the perplexity, the computation starts from the second token rather than including the first generated token. This avoids the potential issue that the initial token is assigned a very low probability and acts as an outlier. Including the initial token could unintentionally correlate the perplexity with the generation length, as its effect diminishes when averaged over a longer sequence.
   
\noindent\textbf{Alignment Adjustment for Qwen2.5-7B Fine-Tuning}. Notably, when applying fine-tuning to Qwen2.5-7B, a challenge is that standard LoRA sometimes failed to achieve proper alignment between the model's generation and the fine-tuning data, particularly when more removal was involved. To address this, we applied a backdoor technique by adding a control phrase to the prompt during fine-tuning. Specifically, we appended "Answer should end with 'The answer is'" at the end of the question in the fine-tuning data. During inference, we included the same phrase to reinforce the pattern learned from fine-tuning, ensuring better alignment in the model's response generation.

\section{Additional related works}

\noindent\textbf{Test-Time Scaling Law.}
{There are some recent discoveries of the test-time scaling law \cite{brown2024large,snell2024scaling,saad2024archon}.}
{While our method focuses on enhancing the reasoning efficiency through removing unimportant reasoning steps from the data, one may question whether this contradicts to the test-time scaling law. To explain this, there is no self-reflection/self-correction mechanism considered in this work, and there is only one reasoning path for each example/fine-tuning data, and we observe an accuracy-token length trade-off. 
In contrast, for test-time scaling law, if we explore more reasoning paths, such an over-thinking can help obtain the correct answer. Our method is perpendicular to the test-time scaling law, and the idea of removing unimportant reasoning steps in our work is also applicable to the test-time methods to reduce the computation cost as well. }

\section{Additional Experiment Details.}\label{sec:appd:hyper}

\noindent\textbf{Hyperparameters in Fine-tuning.}
For SFT, we set the batch size to 128, the learning rate to 5e-5, and the training epoch to 3.0 for all datasets. For ORPO, the batch size is 64, learning rate is 5.0e-6 and training epoch is 5.0. The optimizer for all fine-tuning experiments is AdamW~\cite{loshchilov2019decoupled}.

\newpage
\section{Additional Validation to Support the Design of $t_1$.}\label{sec:appd:t1}
In this section,
we provide additional empirical experiment to demonstrate that when $\text{PPL}_{\text{rem}}$ is larger, it is more necessary to conduct merging.

We manually examine several removal cases, where a reasoning step is eliminated, and categorize them into three classes:

\noindent (1) No coherence issue – Removing the step does not disrupt reasoning, so merging is unnecessary.\\
(2) Minor coherence issue – Removing the step slightly affects coherence; merging is beneficial but not essential. \\
(3) Obvious coherence issue – Removing the step leads to a clear loss of coherence, making merging necessary. \\
For each case, we compute the perplexity change ratio (after removal / before removal) and plot the results in Figure~\ref{fig:coh}. 

\begin{figure}[h]
    \centering
    \includegraphics[width=0.8\linewidth]{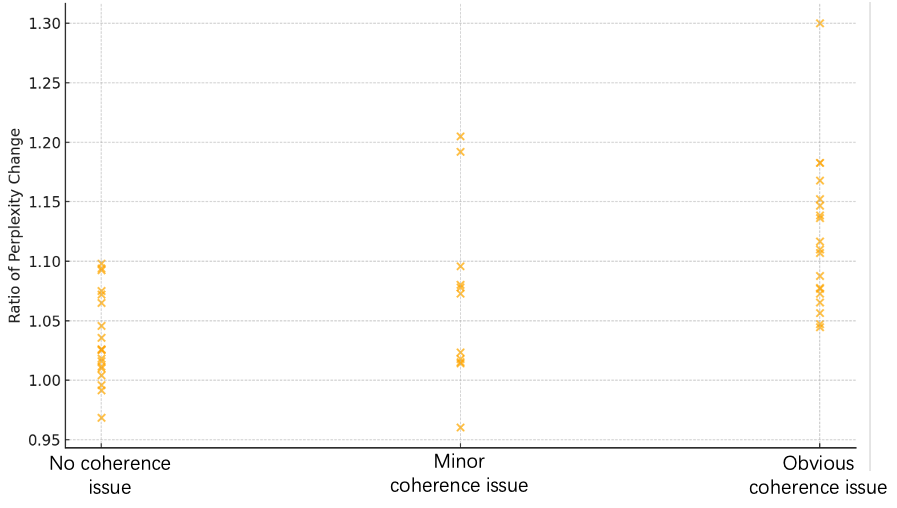}
    \caption{Perplexity Change across different cases.}
    \label{fig:coh}
\end{figure}
From Figure~\ref{fig:coh}, we observe that, in general, the greater the need for merging, the higher the perplexity change. These results support the choice of setting a threshold $t_1$, as when the perplexity gain is small, direct removal is sufficient, making merging unnecessary.
\newpage
\section{Prompt for Instructing Merging}\label{sec:appd:prompt}
\begin{adjustwidth*}{0cm}{-0cm}
\fontsize{9pt}{9pt}\selectfont
\begin{tabular*}{\textwidth}{p{\textwidth}}
\toprule
\midrule
Example 1:\\ 
Q: A portable computer drive has enough kilobytes of storage space to store 2000 photos. Each photo is 1.5 kilobytes in size. How many 200-kilobyte videos can it store if it already has 400 photos on it? Let's think step by step.\\
A: The drive has 2000 photos * 1.5 kilobytes = 3000 kilobytes of storage space.\\
It already has 400 photos * 1.5 kilobytes = 600 kilobytes of photos on it.\\
So, it has 3000 - 600 = 2400 kilobytes of storage space left.\\
Each video is 200 kilobytes in size.\\
The answer is 12
\\ \\
After removing 'The drive has 2000 photos * 1.5 kilobytes = 3000 kilobytes of storage space.', the answer become:\\
A: The drive already has 400 photos * 1.5 kilobytes = 600 kilobytes of photos on it.\\
So, it has 2000 photos * 1.5 kilobytes - 600 kilobytes = 2400 kilobytes of storage space left.\\
Each video is 200 kilobytes in size.\\
The answer is 12\\
\\
Example 2: \\
Q: Leah bought 3 boxes of birdseed. When she went to put them away, she discovered that she already had 5 boxes in the pantry. Her parrot eats 100 grams of seeds each week and her cockatiel eats 50 grams of seeds in a week. If each box of birdseed contains 225 grams, for how many weeks can she feed her birds without going back to the store? Let's think step by step.\\
A: Leah has a total of 3 + 5 = 8 boxes of birdseed.
Each box contains 225 grams of birdseed, so she has a total of 8 * 225 = 1800 grams of birdseed.\\
Her parrot eats 100 grams of seeds each week and her cockatiel eats 50 grams of seeds in a week, so together they eat 100 + 50 = 150 grams of seeds per week.\\
To find out how many weeks the birdseed will last, we can divide the total amount of birdseed (1800 grams) by the amount eaten per week (150 grams):\\
1800 / 150 = 12 weeks\\
The answer is 12\\
\\ \\
After removing 'Leah has a total of 3 + 5 = 8 boxes of birdseed.', the answer become:\\
A: Each box contains 225 grams of birdseed, so Leah has a total of (3 + 5) * 225 = 1800 grams of birdseed.\\
Her parrot eats 100 grams of seeds each week and her cockatiel eats 50 grams of seeds in a week, so together they eat 100 + 50 = 150 grams of seeds per week.\\
To find out how many weeks the birdseed will last, we can divide the total amount of birdseed (1800 grams) by the amount eaten per week (150 grams):
1800 / 150 = 12 weeks\\
The answer is 12\\
\\
Example 3:
Q: Joy can read 8 pages of a book in 20 minutes. How many hours will it take her to read 120 pages? Let's think step by step.\\
A: To find out how many minutes it takes her to read 1 page, we divide 20 minutes by 8 pages: 20 minutes / 8 pages = 2.5 minutes per page.\\
To find out how many minutes it takes her to read 120 pages, we multiply 2.5 minutes per page by 120 pages: 2.5 minutes per page * 120 pages = 300 minutes.\\
To convert minutes to hours, we divide 300 minutes by 60 (since there are 60 minutes in an hour): 300 minutes / 60 = 5 hours.\\
The answer is 5
\\ \\
After removing 'To convert minutes to hours, we divide 300 minutes by 60 (since there are 60 minutes in an hour): 300 minutes / 60 = 5 hours.', the answer become:\\
A: To find out how many minutes it takes her to read 1 page, we divide 20 minutes by 8 pages: 20 minutes / 8 pages = 2.5 minutes per page.
To find out how many minutes it takes her to read 120 pages, we multiply 2.5 minutes per page by 120 pages: 2.5 minutes per page * 120 pages = 300 minutes.\\
The answer is (300 / 60) = 5\\
\\
Learn from the above example to do the following modification. Remember not to change the final results (the number after 'The answer is').\\
\bottomrule
\end{tabular*}

\end{adjustwidth*}

\end{document}